# Automated 3D recovery from very high resolution multi-view satellite images

Rongjun Qin, Assistant Professor

The Ohio State University
Department of Civil, Environmental and Geodetic Engineering
Department of Electrical and Computer Engineering.
218B, Bolz hall, 2036 Neil Avenue, 43210, Columbus, OH, USA.
(+1) 614-292-4356 (phone), qin.324@osu.edu

**ABSTRACT**

This paper presents an automated pipeline for processing multi-view satellite images to 3D digital surface models (DSM). The proposed pipeline performs automated geo-referencing and generates high-quality densely matched point clouds. In particular, a novel approach is developed that fuses multiple depth maps derived by stereo matching to generate high-quality 3D maps. By learning critical configurations of stereo pairs from sample LiDAR data, we rank the image pairs based on the proximity of the results to the sample data. Multiple depth maps derived from individual image pairs are fused with an adaptive 3D median filter that considers the image spectral similarities. We demonstrate that the proposed adaptive median filter generally delivers better results in general as compared to normal median filter, and achieved an accuracy of improvement of 0.36 meters RMSE in the best case. Results and analysis are introduced in detail.

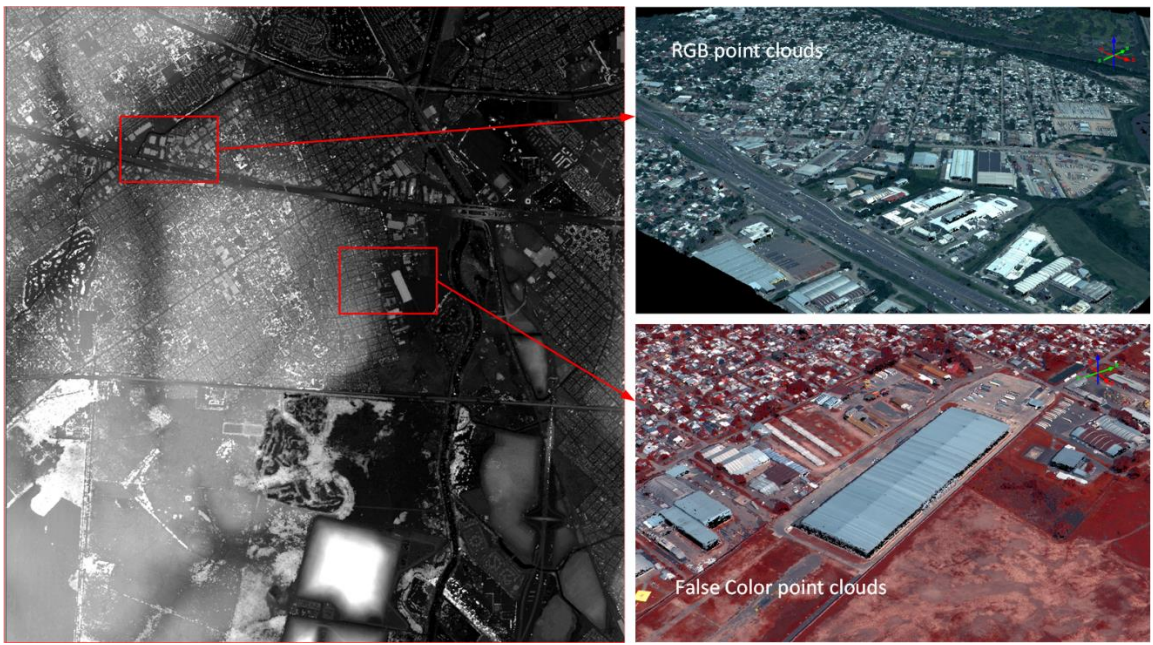

**Overview of 3D recovery from multi-view satellite images**

# INTRODUCTION

The frequent flyovers of the growing number of optical satellite systems collect a vast amount of high-resolution images daily. It is expected that, with the existing and future satellite sensors, any regions of interest in the world, can be viewed by multiple-looks. Rapid and fully automatic 3D recovery from multi-view images can be extremely useful for information extraction and global situational awareness. Although in the past decade there is a great improvement in 3D recovery algorithms from 2D frame images, the use of such big volume multi-view satellite images for large-scale 3D recovery is not well-investigated, partly due to the availability of data and unpopular sensor models to the larger computer vision community. Nowadays almost all the satellite image providers offer alternative parametric models – Rational polynomial functions (RPF), to replace the complicated linear-array sensor model, such that the modelers do not need to worry about implementing different types of camera models and parameter protocols when using images across satellite sensors and platforms.

3D reconstruction is a heavily investigated problem in the field of photogrammetry and computer vision, with more than 50 years of history, and there are tons of literature on techniques and methodologies in solving both orientation problems (geo-referencing, structure from motion, sparse reconstruction) and dense image matching (DIM) problems (or dense correspondence search problem). Both problems have set forth well-developed approaches that could handle ideal or medium quality data, yet in many cases novel and robust approaches are still needed to be developed. While performing binocular stereo-view based 3D reconstruction is still of particular interest (e.g. Robotics, stereo mapping), given the fact that nowadays we are flooded by images and have so many ways of performing digital image acquisition, it is of a great interest to deal with multi-view images as they provide redundant information for empowering the potential of high-quality 3D products.

Commercial and open-source solutions for processing multi-view images, such as those taken by drones with street-view images are readily available and have been adopted for many different types of applications. Although significant improvement is still needed in modeling complex structures, the existing methods have demonstrated substantial improvements over those from ten years before, particularly the dramatically improved point density and ability to match texture-less areas in piece-wise structures. There are quite a few works on stereo matching using satellite stereo images, whereas much fewer on 3D reconstruction from multi-view satellite images. In general, there are three aspects that differentiate the problem of 3D reconstruction of multi-view satellite images to normal multi-view stereo reconstruction:

1. The geometric model: most of the satellite imaging sensors are linear array, being "half" perspective (in the sampling direction) "half" parallel (in the moving direction). Although the geometric model is "reprocessed" to the RPF model, standard operations such as forward projection (for inverse model) and spatial restitution requires iterative solutions.

2. The data quality: Due to the acquisition difference, the viewing angles of the satellite sensors are partially restricted by the orbit, the perspective differences in the moving direction for along-track data. Given that taking one image using linear-array camera requires consistent exposure through time, the number of images along a track is only a few, while off-track data are taken weeks/months apart, under different illuminations and potential physical changes of the ground object. In addition, the radiometric quality of satellite images is usually lower than aerial and drone images due to the degradation of the atmospheric difference. Particularly, objects under the shaded areas are hardly visible and this creates consistent problems. The inconsistence of radiometric and large time differences may potentially create problems in both feature matching and dense reconstruction.

3. Data volume: Satellite data are large in volume; a frame of the satellite image can easily go up hundreds and thousands of mega pixels. Therefore, computational efficiency and memory management are critical factors to test the algorithms.

This paper presents an operational solution for 3D reconstruction based on the multiple views, as a result of the author's participation in a worldwide challenge in 3D multi-view reconstruction (IARPA 3D Multi-view Challenge) (IARPA, 2016). The author presented a solution of multi-depth fusion using pair-wise reconstructed depth map from a selection of pairs. We demonstrate that this solution is computationally efficient and can be used to produce high-quality digital surface models within affordable time frames and computational resources.

## RELATED WORKS

As we described in our previous review work (Qin et al., 2016), the current strategies for performing 3D reconstruction from multi-view images can be coarsely categorized depending on the how images are structured (Qin et al., 2016), with exceptional cases that combine both, being 1) multi-stereo matching (MSM); 2) Multi-view matching (MVM). MSM is a direct extension of two-view stereo matching, in which images are paired and point clouds of each pair are fused/filtered to form a final point cloud (Haala and Rothermel, 2012; Hirschmüller, 2005). MVM considers matching points across multiple images simultaneously (Baltsavias, 1991; Furukawa and Ponce, 2010). MVM is a more rigorous way to incorporate redundant information, but often more complicated to implement. A recent review (Remondino et al., 2014) in DIM compared different software packages (contain methods from both MSM and MVM categories) in generating point clouds from consumer grade images. No specific conclusions were given on the performance of all test methods, due to the complex test cases and flexibility of tunable parameters. Both types of methods have advantages and disadvantages, and their performances vary with the camera network, scene content, and complexity, strategies for point matching (global or local) etc. Our own experience is that generally for top-view photogrammetric images blocks (60-80% overlap for frame images and 15-25 degrees of intersection angle for satellite images), the MSM methods such as SGM (semi-global matching) appear to be a good choice, it leverages both speed and performances (d'Angelo and Reinartz, 2011; Krauß et al., 2013). However, for images taken from terrestrial and mobile platforms, especially for those that form large baselines and poor camera networks, MVM methods in general produce more complete point clouds, since the visibility are modeled while many stereo algorithms tend to resist objects with large parallax (Morgan et al., 2010; Seitz et al., 2006)

In the IARPA multi-view challenge (IARPA, 2016), four out of the top-five finishers end up with a solution that adopts MSM strategies, with the fifth finisher using a partial MVM strategies. It was generally agreed over the past years that the semi-global matching (SGM) – like algorithms provide the best trade-off between the quality and computational efficiency (Gehrke et al., 2010), despite that this algorithm is no longer to the top performer in the computer-vision benchmark lead-board. The top finishers ubiquitously adopt the SGM optimization in their solutions, amongst a variety of different stereo matching algorithms.

As implementing algorithms and developing software even for testing purpose involves a large amount of programming work. Being comparatively less popular than general crowd-source images as in computer vision community, there are still several published works reporting stereo reconstruction systems particularly focusing on satellite images. However, as we pointed out in (Qin, 2016b): the development of light-weight software for RPC stereo images is much less because of the high implementation burden and the smaller size of interested community. There are only a few tools available for DSM generation from RPF modeled images. Most of such either reside in complete commercial software packages or internal institutional developments. Commercial software packages are often well-developed and provide users with easy-to-use interfaces, while reserve less flexibility. For software stability and compatibility reasons, the commercial software packages may not include the latest developments. Some institutions have their internally developed system, which have led to impressive results. One of the examples is the CATENA system developed by (DLR) (Krauß et al., 2013). It implements the classical semi-global matching (SGM) (Hirschmüller, 2008), with complicated consideration in optimization in a distributed system. It is capable of generating country-wide high-resolution DSM fully automatically. SETSM (Noh and Howat, 2015) developed by the BPCRC (Byrd polar and climate research center) in the Ohio State University adopts a TIN-based method for surface extraction, running on a high-performance computer, which has a particularly good performance on glacier areas. Micmac developed by IGN (Deseilligny and Clery, 2011) is an open-source software package for perspective images. According to its manual, it has a small module to generate DSM from RPC images. However, it cannot perform the orientation refinement of the RPC images and has rarely been reported about its processing capability in RPC-based images. Among these works, a few of them mentioned partially strategies of adopting multi-view data to produce a single DSM, while works particularly focusing on Multi-view satellite images are seldom mentioned.

## METHODOLOGY

In this paper, we describe our contribution as a particular multi-depth map fusion while introducing our 3D reconstruction system as a whole. Our reconstruction strategies follow the MSM approach: we first compute pair-wise stereo matching on several stereo pairs and fuse them with the proposed adaptive median filter technique. For stereo matching we perform SGM with a coarse-to-fine strategy to achieve computational efficiency, this is implemented in our stereo matching pipeline RPC Stereo Processor (RSP) (Qin, 2016a). Figure 1 shows a general workflow of the system.

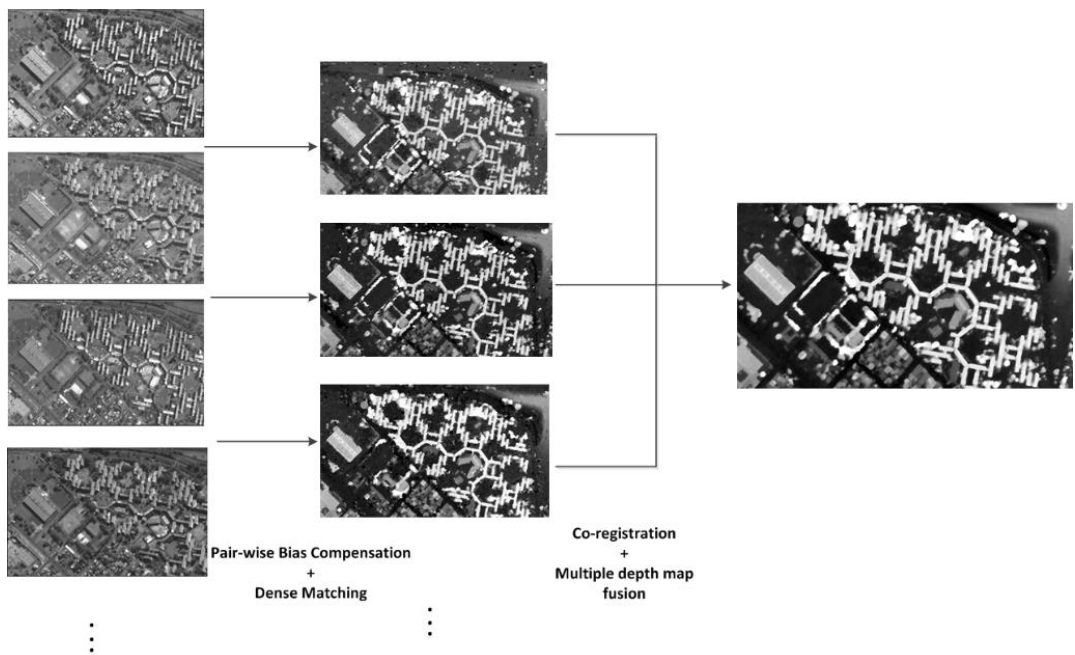

Figure 1. A general workflow of the 3D reconstruction system.

**Global registration**
Given that the RPF provided by the satellite providers have controlled accuracy, as the orientation sensors were frequently calibrated through ground calibration field. Instead of performing a rigorous multi-ray bundle adjustment, we perform a pair-wise bias-compensation using tie points and perform least squares minimization of the generated DSM to a reference DSM, the correction parameters of the registration can be reapplied directly to the RPC (rational polynomial coefficients) given the inverse RPF model using $0^{th}$ order bias-correction:

$$s + \Delta_s = \frac{N_s(U+\Delta U, V+\Delta V, Z+\Delta Z)}{D_s(U+\Delta U, V+\Delta V, Z+\Delta Z)}, \; l + \Delta_l = \frac{N_l(U+\Delta U, V+\Delta V, Z+\Delta Z)}{D_l(U+\Delta U, V+\Delta V, Z+\Delta Z)} \quad (1)$$

Knowing the shift from the DSM registration, we are able to derive the bias-corrector for every image, as an effective bundle adjustment. This idea uses the fact that there are ignorable rotational differences (Waser et al., 2008) between pairs of DSMs and eventually this provide an effective solution for global registration, since the RPC parameters are calculated in a way that there are very few degrees of freedom, which is likely, a systematic shift. Pair-wise relative orientation is performed through the RSP pipeline developed by the author. Interested readers can refer to (Qin, 2016b).

**Pair selection**
It is intuitive to compute all the possible pairs, while the computation time goes exponentially as the number of images increases. Therefore, it is important to choose "good pairs" to compute the depth maps for fusion. As suggested by d'Angelo et al (d'Angelo et al., 2014), one critical factor for choosing the pairs is to limit the intersection angle within a certain range to maximal the results of the SGM algorithm, and in their work the suggested the intersection angle to be within 15-25 degrees. However, we observe this empirical value is effective to only to a certain degree: there are many pairs with intersection angle falling into this range, while rendering poorer results than pairs falling out of this range. We also observe that if the intersection angle is smaller than 8 degree or larger than 40, the results are generally poor, while pairs within 8-40 degrees are still at a large quantity. Choosing the correct pairs is a complicated multi-factor issue. In our case, we adopt a learning approach as we have a small patch of ground truth LiDAR data. The idea of choosing the pairs is simple: we generate DSM only for a small area with a combination of pairs that matches our ground truth file. These pairs are limited to those only with intersection angle between 10-30 degrees. We then rank the pairs after a least squares minimization between the DSM generated from the image pairs to the ground truth of the data. The first ten pairs that generate good results are used for fusion.

**Adaptive Depth Fusion**

An effective method for multiple depth fusion adopts a median filter along the vertical direction. This assumes a stack of 2.5D grid, and median value of the height in each cell of the overlapping grid is assigned as the value of the final DSM. This is the most commonly used strategy for fusing the depth maps. However, a critical issue of this method is that it might create potential noises in flat regions (as shown in Figure 2c), as each cell are filtered independently from each other. Moreover, when fusing different depth map, the color/intensity information of the image are not utilized, while it provides a rich amount of information on the spatial consistency/smoothness. We therefore proposed an adaptive method that considers the spatial consistency when performing the filtering, being adaptive filter: instead of assigning the median value of the height in each individual cell, we define a window centered at this cell, and the candidate height values for filtering will be in those cells within the window. This can dramatic improve the robustness of the median filtering. A potential problem is that a normal rectangle or square window will lead to blurry effects in the boundaries. Inspired by the idea of bilateral filter (Tomasi and Manduchi, 1998), where it considers filtering with points that only have a certain degree of similarity to the centric pixel, by that shape boundaries of the image can be well preserved. A general hypothesis is that the image boundaries to some degree present possible depth discontinuities, which has been successfully utilized by many powerful matching algorithms (Hirschmüller, 2005; Kolmogorov and Zabih, 2001; Yoon and Kweon, 2006). We follow this hypothesis by creating a weighted Gaussian distribution based on the spatial and color proximities to the centric pixel:

$$W(\boldsymbol{x}) = e^{-\frac{||x-x_0||^2}{2\delta_S^2} - \frac{||I-I_0||^2}{2\delta_I^2}} \qquad (2)$$

where $\boldsymbol{x}$ and $\boldsymbol{x_0}$ are the positions of the current pixel and centric pixel, and $\boldsymbol{I}$ and $\boldsymbol{I_0}$ are the intensity or color values of the current and the centric pixel. Computed weights for each window can be normalized to [0, 1]. Given a threshold $\gamma$, we can define an irregular window AW around the pixel to be filtered (Figure 2(a-b)):

$$AW = \{\boldsymbol{x} | W(\boldsymbol{x}) > \gamma\} \qquad (3)$$

A median filter is then applied to all the pixels within this window. The advantage is that it potentially increases the number of possible candidates for statistically more robust median sampling, particular when the number of filter images is small (only a few, e.g. 2-3 images), which is normally the case, such as ZY-3 three-line scanner, Pleiades tri-stereo. Moreover, it fully utilizes the image information associated with the DSM. In addition, this simple formulation will allow us to process large-format images very efficiently. We find that the fused DSM using our method visually appears smoother in flat regions, while keeping sharpen at the depth boundary, and the DSM fused using normal median filter presents a certain degree of salt-and-pepper noise. Experimental statistics is reported in section IV. Figure 2 (c-d) demonstrates the comparison between the DSM filtered by the proposed approach and a normal median filter:

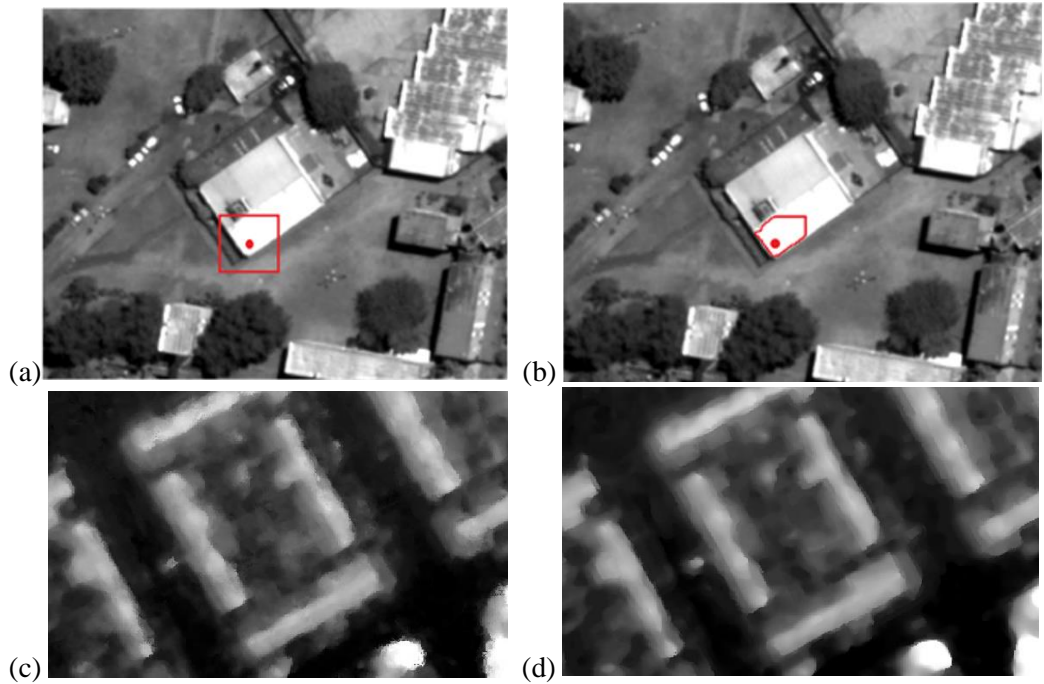

Figure 2. Illustration of the proposed adaptive median filter fusion method: (a) normal rectangular window centered at a pixel. (b) Irregular window (adaptive window) by applying a threshold (0.5) to the weight Gaussian distribution formulated in equation (2); (c) fused DSM with normal median filter; (b) fused DSM with the proposed adaptive median filter.

## EXPERIMENT RESULTS

**Test regions**
The system presented here has been tested using John's Hopkins University Applied Physics Lab's (JHUAPL) benchmark dataset (Bosch et al., 2016), where 50 worldview2/3 images are acquired over the same region (near San Fernando) across two years. These images are taken under various conditions containing on-track and off-track stereos, as well as images with seasonal differences. The readers may refer to (Bosch et al., 2016) for more details about the data. It is a challenge to process such into three-dimensional point clouds. Given that the LiDAR reference data is available in this region, it makes possible to perform the accuracy analysis of the proposed method. We choose two sub-regions (ca. 1 km$^2$) that were used for the IARPA multi-view 3D contest, the region is shown in Figure 3. Both sub-regions contain a mixture of high-rise and low-rise buildings, in particular the test region 2, there are a few buildings are around 30 meters high to the ground. This will create large parallaxes on the stereo images and lead to matching failures.

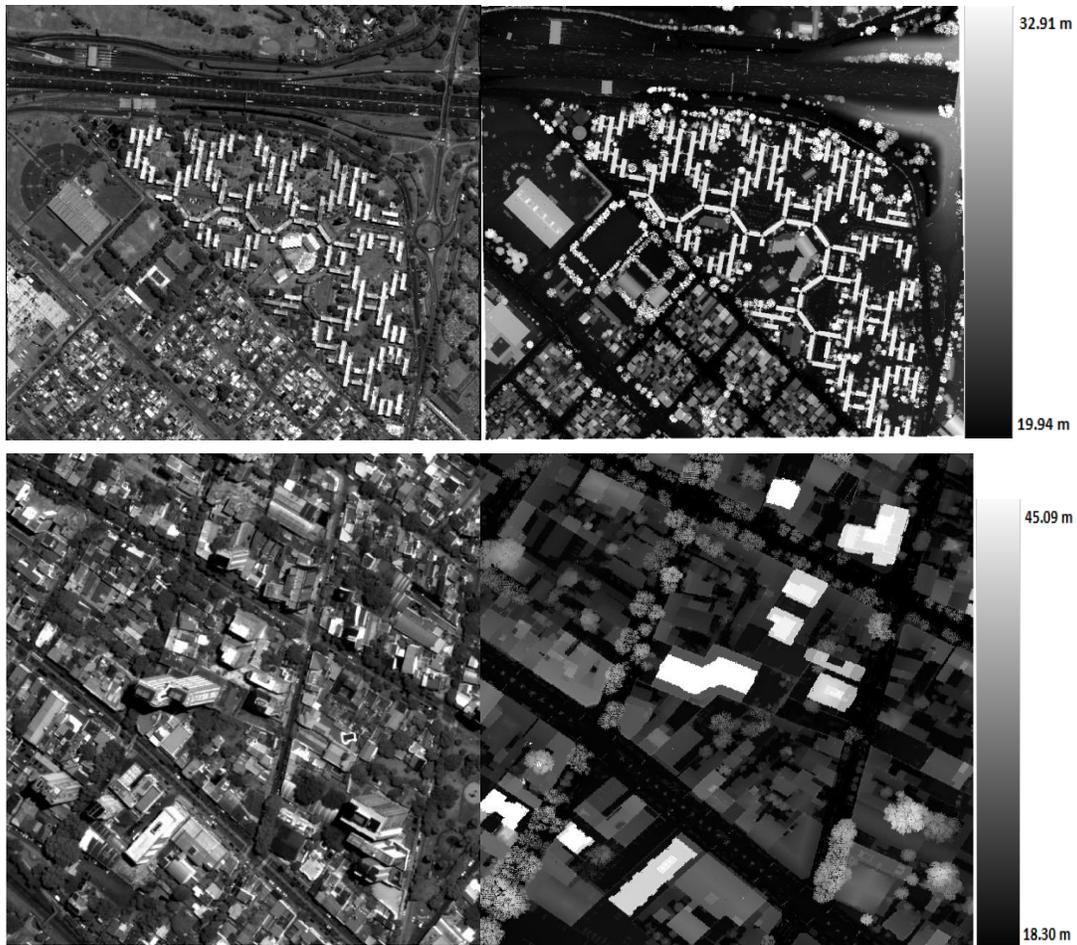

Figure 3: row 1: test region 1; row 2, test region two. Left figures: the panchromatic images; right figures: the ground truth LiDAR data.

**Evaluation**

We follow a rigorous approach in evaluating the resultant DSM from the images to the ground truth LiDAR data, as there might be systematic shifts between the generated DSM from due to systematic errors of the orientation files. We first align both DSMs (the computed DSM and the LiDAR-derived DSM) using a least squares minimization method, by computing the root mean square errors (RMSE) of the height difference. In the minimization process, to eliminate the blunders (such as seasonal differences between the two DSMs), we did not consider points that whose height is larger than a certain threshold (in our case we use 6 meters). Once the datasets are fully aligned, we compute the RMSE of all the points (including those blunders). Figure 4 shows the results of our proposed method when five stereo pairs are used for fusion. The computed RMSE of the both DSMs are shown in Table 1. It shows in the best case the proposed method achieved 0.36 meter of accuracy improvement (Test region 2, #image=5).We plotted figures showing the performance of the median filter and the proposed adaptive filter in Figure 5. It shows that for most of the cases, the RMSE is smaller than the normal median filter, and there seems to an optimal number of images to be fused that leads to the smallest errors, which is around 4-5 images for the proposed method, consistent for both test regions.  The test region 1 shows that for the normal median filter obtained the minimal RMSE when the number of images is three, while test 2 shows that the more image it incorporates the better results it delivered.

However, we understand this is not a general conclusion as the results of the evaluation may be biased by the seasonal differences and quality of pairs at different regions.

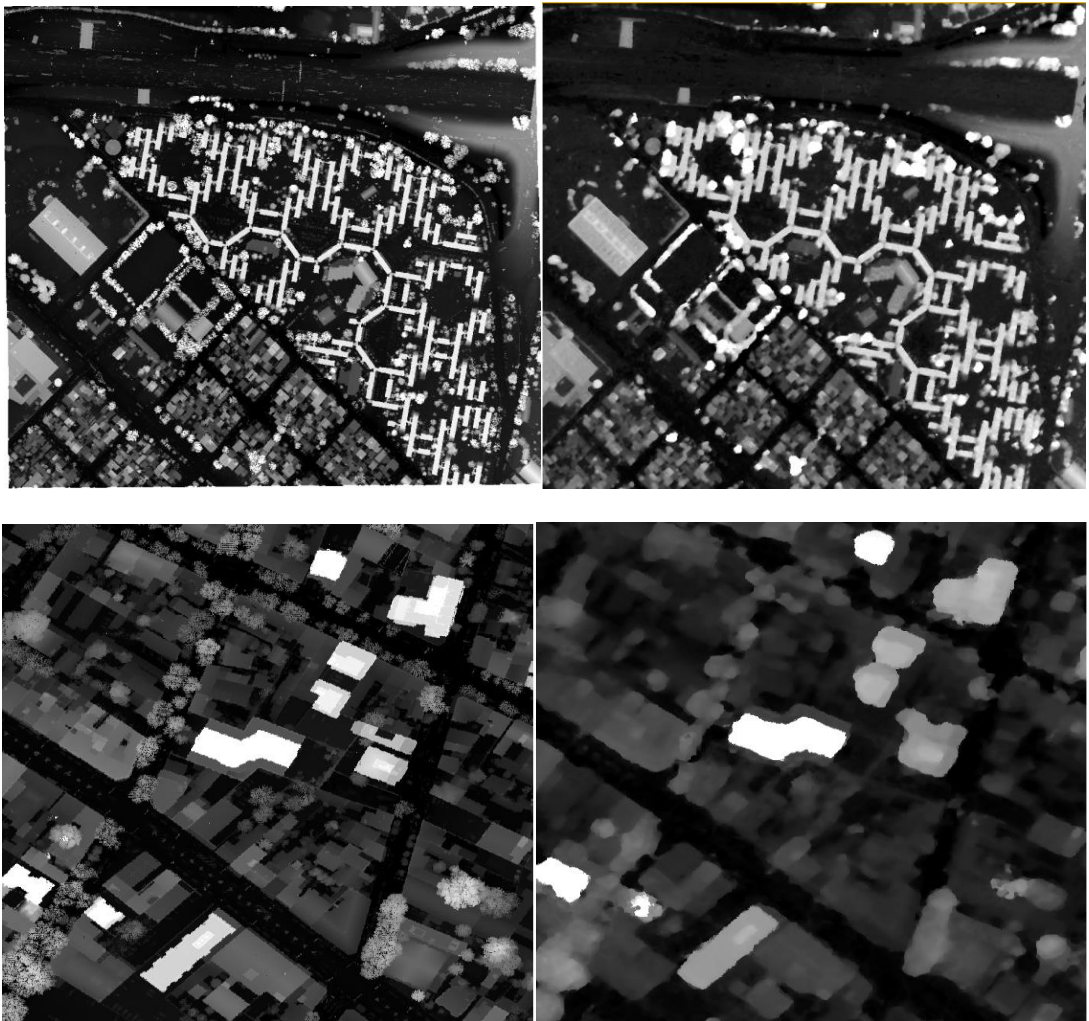

Figure 4: Visual results of the computed DSM versus the ground-truth LiDAR data. Row 1: test region 1; row 2, test region two. Left figures: the ground-truth LiDAR Data; right figures: the computed DSM using our proposed method.

Table 1. RMSE between the generated DSM and the Ground truth LiDAR data (meter)

| Test Region | Method | # image=1 | # image=2 | # image=5 | # image=10 |
|---|---|---|---|---|---|
| Region 1 | Proposed method | 2.617 | 2.528 | 2.513 | 2.541 |
| Region 1 | Median Filter | 2.620 | 2.687 | 2.562 | 2.563 |
| Region 2 | Proposed method | 5.429 | 4.816 | 4.530 | 4.556 |
| Region 2 | Median Filter | 5.146 | 4.951 | 4.836 | 4.628 |

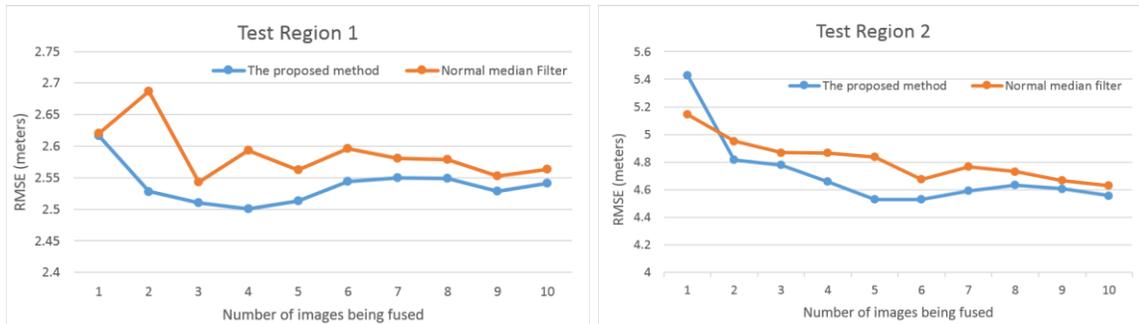

Figure 5 Error analysis of both test regions on the RMSE and number of images being fused.

**Conclusion**

We present a simple solution for fusing multiple depth maps by utilizing the color information of the orthophoto. The results of the experiment on JHUAPL have shown that our proposed methods delivers better results in general, and achieved an accuracy of improvement of 0.36 meters RMSE in the best case. The results also show that the proposed method generally obtain the best results when fusing about 4-5 images. This is because once more DSMs generated from low-quality pairs are involved; errors will be propagated through the window, while normal median filter has lighter problems in this case. It also shows that the proposed method is able to maintain the spatial smoothness of the DSM, as compared to normal median filter, which generates salt-and-pepper errors. This is a small pffigurart of a larger system the author develops to handle wide-area 3D reconstruction, where we are able to produce high-precision multi-view satellite images of 36 square $km^2$ within 6.5 hours in single PC. For more details and trial of our system, the readers are encouraged to contact the authors.

**Acknowledgements:**

The author would like to thank John Hopkins University Applied Physics Lab to support the Imagery, and the author would also like to thank the IARPA to organize the 3D challenge available that drives forth this work.